# Rough Clustering Based Unsupervised Image Change Detection

Chandranath Adak
Department of Computer Science and Engineering
University of Kalyani
West Bengal-741235, India
mr.c.adak@ieee.org

*Abstract*—This paper introduces an unsupervised technique to detect the changed region of multitemporal images on a same reference plane with the help of rough clustering. The proposed technique is a soft-computing approach, based on the concept of rough set with rough clustering and Pawlak's accuracy. It is less noisy and avoids pre-deterministic knowledge about the distribution of the changed and unchanged regions. To show the effectiveness, the proposed technique is compared with some other approaches.

## I. INTRODUCTION

*Change detection* [1-2] is a technique to identify and analyze the differences of images/objects on the same reference plane at different times. It can be utilized in remote sensing [3-4], underwater sensing [5], weapon detection, video surveillance [6], medical diagnosis and treatment [7], civil and mechanical infrastructure [8], driver assistance systems [9], vegetation changes [10], urban growth and shifting cultivations monitoring [11] etc.

In change detection, there may be two approaches: supervised and unsupervised [12]. In the maximum cases, unsupervised techniques are more appropriate than supervised ones due to unavailability of platform truth information. The change detection can be observed as a clustering [13] process, i.e. classification of the data into two sets *changed* and *unchanged*. The patterns, 'within the cluster' and 'between the clusters' are *homogeneous* and *heterogeneous* respectively [14].

Here rough clustering [15-16, 22] is used with rough set, where lower approximation represents unchanged clusters and upper approximation represents the changed clusters. Pawlak's accuracy [17] is used to generate difference image. Our proposed method is compared with *Hard C-Means* (HCM), *Fuzzy C-Means* (FCM), methods reported in [20-21, 14].

## II. ROUGH CLUSTERING

*Rough clustering* (Prado, Engel, Filho, 2002; Voges, Pope, Brown, 2002) is an expansion work of rough (approximation) sets, which is pioneered by Pawlak (1982, 1991).

### A. Information System Framework

In Rough set theory, an assumption is granted, i.e. information is related with each and every entry of the data matrix. The over-all information expresses the completely available object-knowledge. More precisely, the information system is a pair of tuples, $S=(U,A)$, where U is a non-empty finite object set called as universe and $A=\{a_1,...,a_j\}$ is a non-empty finite attribute set on U. With every attribute $a \in A$, a set $V_a$ is allied such that $a : U \rightarrow V_a$. The set $V_a$ is called the domain (value) set of *a*.

### B. Equivalence Relation

An associated *equivalence relation* resides with any $P \subseteq A$,

$$IND(P)=\{(x,y) \in U^2 \mid \forall a \in P, a(x)=a(y)\}.$$

The equivalence relation *IND(P)* is termed as a *P-indiscernibility* relation. The partition of U is a family of all *equivalence classes of IND(P)*, denoted by *U/IND(P)* or *U/P*. Those *x* and *y* are *indiscernible* attributes from *P*, when $(x,y) \in IND(P)$.

### C. Rough Set

The main thing of rough set is the equivalence between objects (known as indiscernibility). The equivalence relation is formed with same knowledge-based objects of the information system. The partitions (formed by division of equivalence relations) build the new subsets. An information system $S=(U,A)$ is assumed, such that $P \subseteq A$ and $X \subseteq U$. The subset X (using information contained in attributes from P) is described by constructing two subsets: *P-lower* approximations of X $(P_*(X))$ and *P-upper* approximations of X $(P^*(X))$, where:

$$P_*(X) = \{ x \mid [x]_P \subseteq X \} \text{ and } P^*(X) = \{ x \mid [x]_P \cap X \neq \emptyset \}.$$

Sometimes, an additional set $(P_D(X))$, i.e. the difference between the upper approximation $(P^*(X))$ and the lower approximation $(P_*(X))$ becomes very effective in analysis.

$$P_D(X) = P^*(X) - P_*(X)$$

The *accuracy* $(\alpha_P)$ of the rough-set (Pawlak, 1991) illustration of the set X is as follows:

$$0 \leq ( \alpha_P(X) = |P_*(X)| / |P^*(X)| ) \leq 1$$

Rough clustering is the expansion of rough sets, containing two extra requirements: an ordered attribute value set and a distance measurement (Voges, Pope & Brown, 2002). As like standard clustering techniques, distance measurement is done by ordering value set, and clusters are generated by these distance measure.

### III. PROPOSED TECHNIQUE

The steps of the proposed technique as follows:

Step 1: Computational overhead is reduced by transformation of each RGB image pixel value into a single valued attribute:

$$Pixel_T = Pixel_R + 2*Pixel_G + 3*Pixel_B$$

Step 2: Single valued transformed image pixel values ($Pixel_T$) are the input data set $X = \{x_1, x_2, ..., x_n\}$.

Step 3: *P-lower* and *P-upper* approximations of $X$ are calculated, $P_*(X) = \{ x \mid [x]_P \subseteq X \}$ and $P^*(X) = \{ x \mid [x]_P \cap X \neq \emptyset \}$; $P_*(X)$ is the *unchanged clusters* and $P^*(X)$ is the *changed clusters* with respect to the reference image.

Step 4: The *accuracy* ($\alpha_P$) is calculated by:

$$\alpha_P(X) = |P_*(X)| / |P^*(X)|$$

Step 5: All the pixels ($Pixel_{DI}$) are chosen for which $\alpha_P(X) \geq T$. $T$ is a threshold [18] value ($0 \leq T \leq 1$).

Step 6: Difference Image (DI) is generated by $Pixel_{DI}$.

[ *used convention* : changed pixel color = white, unchanged pixel color = black ].

### IV. EXPERIMENTAL RESULTS AND COMPARISON

To assess the immovability and accurateness of the proposed technique, the results are obtained from different images.

*Fig.1* (satellite image, *courtesy:* «Landsat» program, since 23 July 1972, run jointly by NASA & U.S. Geological Survey) shows a specific region of Saudi Arabia viewed as desert in 1987. But by 2012, a lot of the area converted into green agricultural land [19].

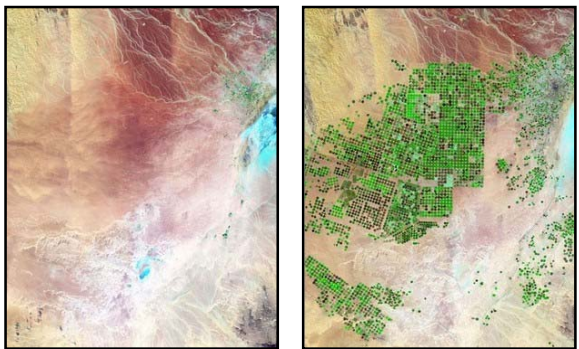

*Fig. 1.(a)*  *Fig. 1.(b)*

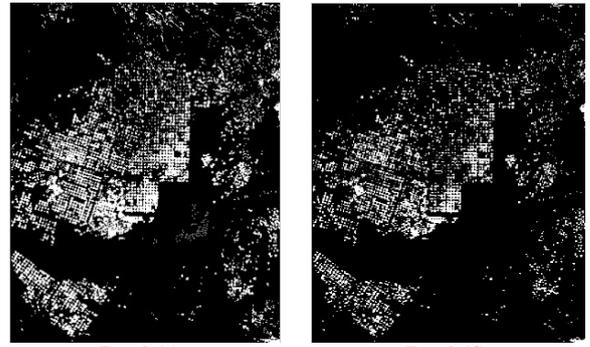

*Fig. 1.(c)*  *Fig. 1.(d)*

*Fig.1.* Satellite image of Saudi Arabia geographical region: *(a)* acquired in 1987, *(b)* acquired in 2012, *(c)-(d)* changed region detected by FCM and proposed rough clustering based technique ( *T=0.5* ) respectively.

*Fig.2* shows the effectiveness of the proposed technique in medical image processing (e.g. cell-patch detection).

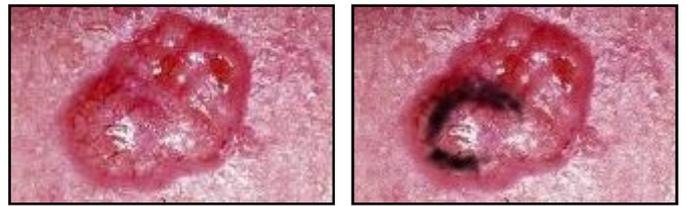

*Fig. 2.(a)*  *Fig. 2.(b)*

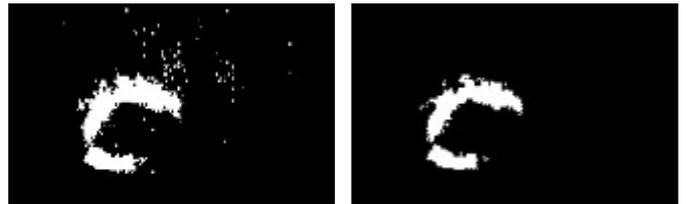

*Fig. 2.(c)*  *Fig. 2.(d)*

*Fig.2.* Cell image: *(a)* without patch, *(b)* with patch, *(c)-(d)* change detection by HCM and proposed technique ( *T=0.55* ) respectively.

*Fig.3* assesses the proposed technique with respect to video surveillance (e.g. hall monitoring).

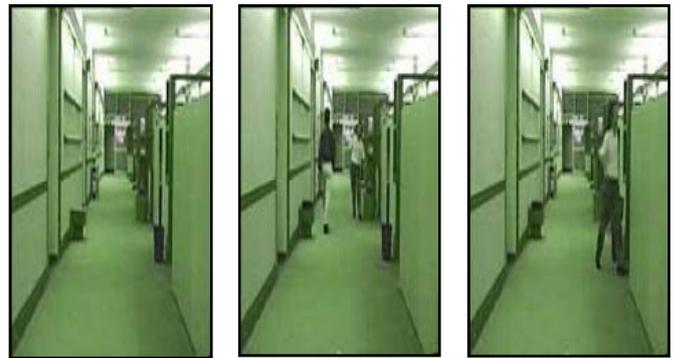

*Fig. 3.(a)*  *Fig. 3.(b)*  *Fig. 3.(c)*

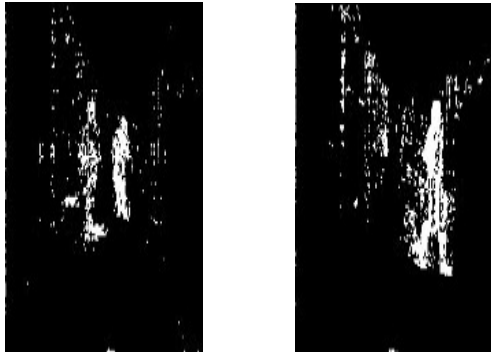

Fig. 3.(d)    Fig. 3.(e)

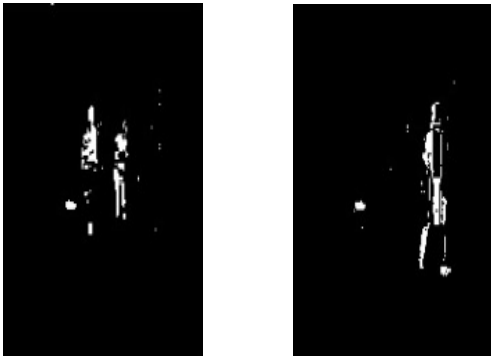

Fig. 3.(f)    Fig. 3.(g)

*Fig.3.* Hall monitoring in QCIF (176x144), 30fps, YUV 4:2:0 video [21]: *(a)* reference frame, *(b)-(c)* frames at different time intervals, *(d)-(e)* change detection of (b)&(c) with reference frame(a) by IOM [20-21], *(f)-(g)* change detection of (b)&(c) with reference frame(a) by proposed rough clustering based technique (*T=0.52*).

More testing outputs are shown in *fig.4-5*.

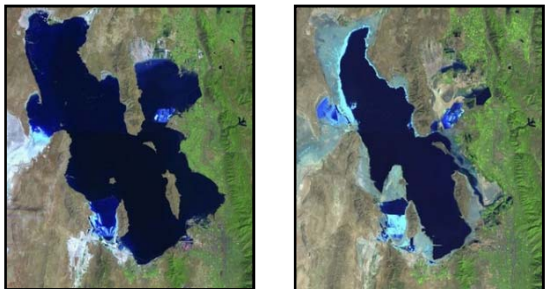

Fig. 4.(a)    Fig. 4.(b)

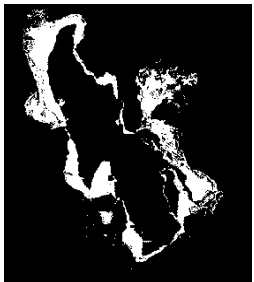

Fig. 4.(c)

*Fig.4.* Satellite image [19]: *(a)* previously acquired region, *(b)* recently acquired region, *(c)* change detection by proposed rough clustering based technique ( *T=0.3* ) .

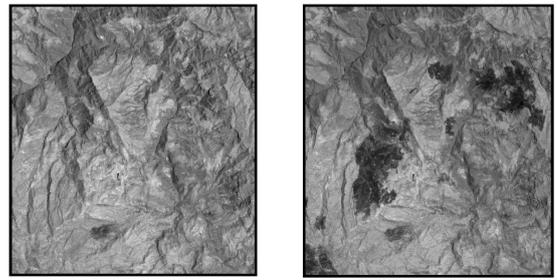

Fig. 5.(a)    Fig. 5.(b)

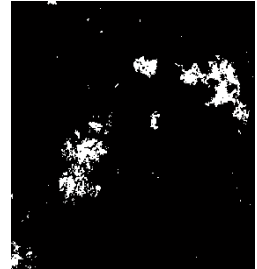

Fig. 5.(c)

*Fig.5.* Mexico area - Band 4 image, comparisons reported in [14]: *(a)* acquired in 18th April 2000, *(b)* acquired in 20th May 2002, *(c)* change detection by proposed method(*T=0.5*).

From the above test results (*Fig. 1-5*), it is easily observed that the output of the proposed technique is less noisy, more precisely clustered and stable.

## V. CONCLUSION AND FUTURE WORK

The proposed method is tested on different images (i.e. *satellite image*; *medical image*: cell patch, breast cancer patch, bone fracture in X-Ray image, MRI image, teeth cavity image; *video frames*; *remotely sensed image*: vegetation changes, civil infrastructure, burned area assessment, urban growth image etc.). It produces stable, more noiseless and fairly good results in every case, which assesses the high robustness of this technique. The performance of the proposed technique is compared with some other algorithms and it works fine. The limitation of this type of change detection technique is that the multitemporal images are dependent on the same static reference plane. The next venture of this prescribed technique is to combine rough clustering with different fuzzy clustering and classifiers [23].

### ACKNOWLEDGMENT

I would like to heartily thank *Prof. Bidyut B. Chaudhuri*, IEEE Fellow, Head, Computer Vision and Pattern Recognition Unit, Indian Statistical Institute, Kolkata 700108, India for discussion various aspects in this research field.